%% file: main.tex
\documentclass[letterpaper, 10 pt, conference]{ieeeconf} 

\IEEEoverridecommandlockouts 

\usepackage{comment}
\usepackage[dvipsnames]{xcolor}
\usepackage[hidelinks]{hyperref}
\usepackage{graphicx}

\definecolor{myPink}{RGB}{223,5,180}

\newcommand{\dbname}{Med-RAD}

\title{\LARGE \bf
A Dataset of Anatomical Environments for Medical Robots: Modeling Respiratory Deformation
}

\author{Inbar Fried$^{1,2}$, Janine Hoelscher$^{1}$, Jason A. Akulian$^{3}$, and Ron Alterovitz$^{1}$
\thanks{This research was supported by the U.S. National Institutes of Health (NIH) under awards R01EB024864 and F30CA265234, and by the National Science Foundation (NSF) under award 2008475.}
\thanks{$^{1}$ I. Fried, J. Hoelscher, and R. Alterovitz are with the Department of Computer Science, University of North Carolina at Chapel Hill, Chapel Hill, NC 27599, USA. {\tt\footnotesize \{ifried01, jhoelsch, ron\}@cs.unc.edu}}
\thanks{$^{2}$ I. Fried is also with the Medical Scientist Training Program, University of North Carolina School of Medicine, Chapel Hill, NC, 27599, USA.}
\thanks{$^{3}$J. A. Akulian is with the Division of Pulmonary Diseases and Critical Care Medicine at the University of North Carolina at Chapel Hill, NC 27599, USA. {\tt\footnotesize jason\_akulian@med.unc.edu}}%
}

\begin{document}

\maketitle
\thispagestyle{empty}
\pagestyle{empty}

\input{Text/0_Abstract}
\input{Text/1_Introduction}
\input{Text/2_Methods}
\input{Text/2a_ClinicalScenario}

\input{Text/2b_LungEnvironment}

\input{Text/2c_Deformation}
\input{Text/3_Discussion}

\input{Text/4_acknowledgements}

\bibliographystyle{IEEEtran}
\bibliography{references, bibabbrevs}

\end{document}

%% file: Text/0_Abstract.tex
\begin{abstract}

Anatomical models of a medical robot's environment can significantly help guide design and development of a new robotic system.
These models can be used for benchmarking motion planning algorithms, evaluating controllers, optimizing mechanical design choices, simulating procedures, and even as resources for data generation.
Currently, the time-consuming task of generating these environments is repeatedly performed by individual research groups and rarely shared broadly.
This not only leads to redundant efforts, but also makes it challenging to compare systems and algorithms accurately.
In this work, we present a collection of clinically-relevant anatomical environments for medical robots operating in the lungs.
Since anatomical deformation is a fundamental challenge for medical robots operating in the lungs, we describe a way to model respiratory deformation in these environments using patient-derived data.
We share the environments and deformation data publicly by adding them to the Medical Robotics Anatomical Dataset (\dbname), our public dataset of anatomical environments for medical robots.

\end{abstract}

%% file: Text/1_Introduction.tex
\section{Introduction}

The field of medical robotics is growing rapidly with new robots being actively developed in both industry and academia~\cite{dupont2021decade, attanasio2021autonomy, troccaz2019frontiers}.
Design and evaluation of these systems and their many building blocks, which include motion planners, controllers, mechanical components, user interfaces, and more, are often done first in simulation before progressing to phantom, ex vivo, in vivo, and ideally human trials.
In the initial stages that occur in simulation, accurate and clinically relevant anatomical scenarios are important for proper assessment of the systems and their potential future clinical applicability.
However, despite the extensive amount of research and development in medical robotics, the task of generating these anatomical evaluation environments is repeatedly and redundantly performed by individual research groups.
This is both a time consuming endeavour and creates a lack of standardization in the field which can make it difficult to accurately compare systems and algorithms and reproduce results.

A shared dataset of anatomical environments derived from real clinical cases would eliminate these drawbacks and provide a valuable resource for both model-based and data-driven design and development of medical robots.
In prior work, we presented~\dbname~(formerly named Med-MPD) as a benchmarking dataset with a focus on evaluating motion planners for medical continuum robots~\cite{fried2022clinical}.
The dataset consisted of clinical scenarios in the brain, liver, and lungs with anatomical segmentations and clinically motivated start poses and target positions.
A key missing feature in the dataset was the ability to model anatomical deformation.

Unlike for robots operating in structured environments such as factories or warehouses, medical robots need to operate in deforming, heterogeneous, and uncertainty-ridden environments.
These properties of the environment introduce challenges to both tele-operated robots as well as robots attempting to complete portions of medical procedures autonomously.
Along these lines, in this paper, we extend our prior work by introducing functionality for simulating respiratory deformation in the lungs.
We add three new lung environments to the dataset, including the deformation fields representing respiratory deformation in each environment.
These deformations are extracted from each patient's inspiratory-expiratory CT scan pairs.
We interpolate over the deformation field to simulate the respiratory deformation on the lung anatomy.
The dataset is publicly available at~\url{https://github.com/UNC-Robotics/Med-RAD}.

%% file: Text/2_Methods.tex
\section{Methods}

We briefly describe an example of a clinical scenario for medical robots operating in the lungs.
We then describe the anatomical environment of the lungs, including how we calculate and simulate respiratory deformation.

\begin{figure}[t!]
    \includegraphics[width=0.90\columnwidth]{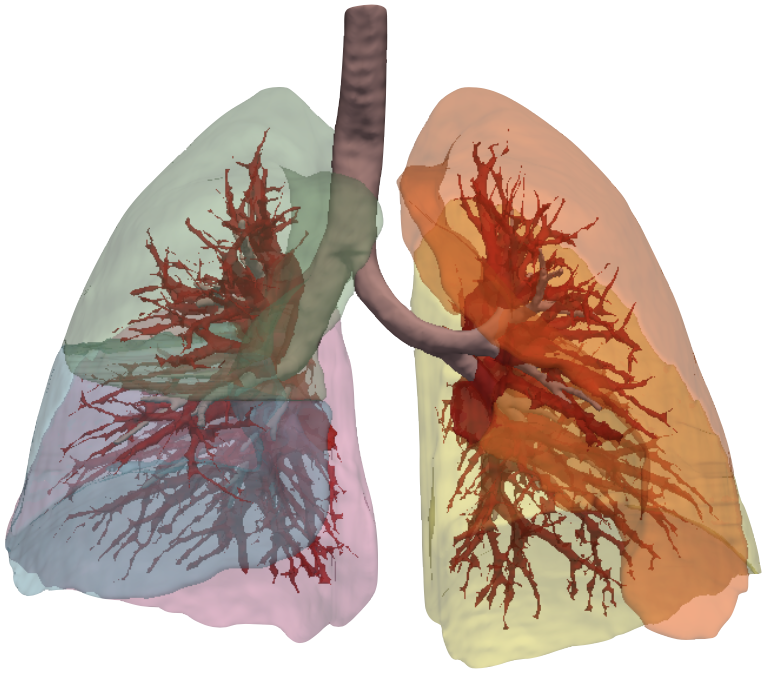}
    \centering
    \caption{
    A representative segmentation of the lungs consisting of the bronchial tree (tan), blood vessels (red), and lung lobes (green, blue, pink, orange, and yellow).
    \vspace{-4mm}
    }
    \label{fig:figLungEnv}
\end{figure}

%% file: Text/2a_ClinicalScenario.tex
\subsection{Clinical Task}

Lung cancer is the leading cause of cancer related deaths in the United States, accounting for over 120,000 deaths annually~\cite{siegel2023cancer}.
When a patient presents with a lung nodule concerning for cancer, a tissue sample is required to establish a definitive diagnosis and guide treatment.
Currently, the safest and least invasive method for acquiring a tissue sample is bronchoscopy where a physician navigates a bronchoscope through the airways and inserts a needle through the airway wall into the lung parenchyma --- the functional tissue of the lungs --- towards the target nodule.
In recent years, there have been significant advances in bronchoscopy that are enabling access to previously hard-to-reach peripheral lung regions using robotic bronchoscopes~\cite{chen2021robotic, yarmus2020prospective, diddams2023robotic, pittiglio2023personalized} and robotic steerable needles~\cite{kuntz2023autonomous}.
However, a major challenge in bronchoscopic biopsy remains overcoming deformation of the lungs from sources such as respiratory motion. 
These deformations can cause large changes in the anatomy, greatly reducing the accuracy of the predicted location of the bronchoscope with respect to the target nodule as well as affecting needle trajectories~\cite{furukawa_comparing_2018, chen_effect_2015}.

%% file: Text/2b_LungEnvironment.tex
\subsection{Anatomical Environment}

The primary anatomical structures in the lungs include the bronchial tree which provides a passageway for air in and out of the lungs, blood vessels, lung fissures which separate lung lobes, and the pleura (i.e., the lung boundary).
We use 3D Slicer~\cite{kikinis_3d_2014} to segment these structures from a given patient's CT scan and use the resulting three-dimensional binary images to represent the anatomical environment.
The value of each voxel in these binary images determines whether the voxel corresponds to an anatomical structure, such as the airways or blood vessels, or to free space. 
We present three environments in the lungs extracted from real patients.
Two of the scans are from the EXACT09 dataset~\cite{lo_extraction_2012} and one scan is from the CT-vs-PET-Ventilation-Imaging study~\cite{eslick2018ct, Eslick_Kipritidis_Gradinscak_Stevens_Bailey_Harris_Booth_Keall_2022} from The Cancer Imaging Archive (TCIA)~\cite{clark2013cancer}.
We selected patients from these datasets which had both an inspiratory and expiratory CT scan.
We use the expiratory CT scan of each patient for segmentation since the intraoperative state of the anatomy is more similar to this state compared to the inspiratory state~\cite{furukawa_comparing_2018, chen_effect_2015}.
The bronchial tree segmentations in these environments include airway generations ranging from the eighth to the twelfth generation.
A representative segmentation is shown in Fig.~\ref{fig:figLungEnv}.

\begin{figure}[b!]
    \includegraphics[width=0.95\columnwidth]{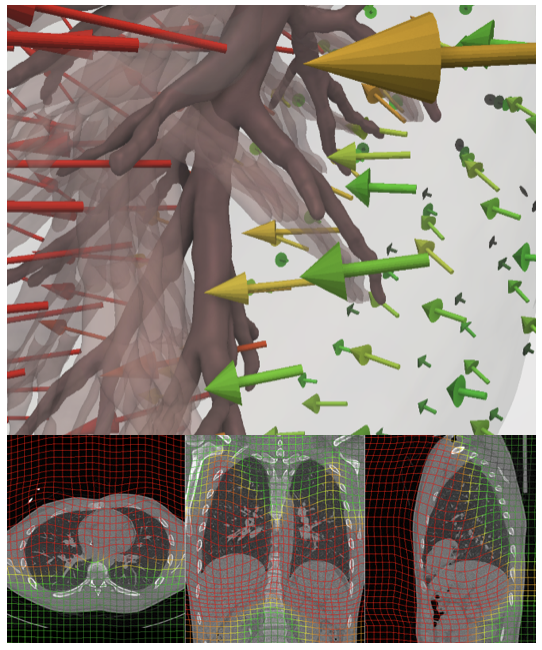}
    \centering
    \caption{
        The result of the deformable image-to-image registration is a three-dimensional vector field indicating the magnitude and direction of displacement across the lungs.
    }
    \label{fig:figRespDef}
\end{figure}

\begin{figure*}[ht]
    \centering
    \includegraphics[width=1.9\columnwidth]{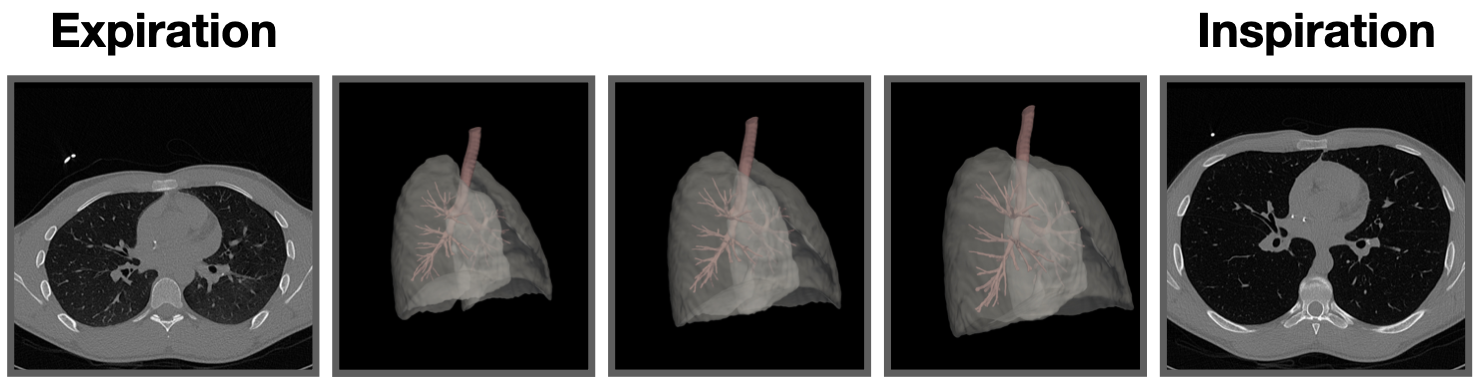}
    \caption{
        We use the expiratory and inspiratory CT scans for each patient to calculate a deformation field. We then interpolate along this field to simulate respiratory motion in the lung environment.
    }
    \vspace{-4mm}
    \label{fig:resp_interp}
\end{figure*}

%% file: Text/2c_Deformation.tex
\subsection{Respiratory Deformation}

Since the patients we selected have both an inspiratory and an expiratory CT scan, we can use these pairings to perform a non-rigid image-to-image registration between these two extreme states of lung anatomy~\cite{klein2009elastix}.
To do so, we first crop both images to remove any regions outside of the lung boundary.
We then use Elastix non-rigid registration within 3D Slicer which uses a b-spline transform to compute the deformation field between the two images for each patient.
The resulting vector deformation field represents the magnitude and direction of deformation across the lungs (Fig.~\ref{fig:figRespDef}).
By interpolating over the deformation field (see Fig.~\ref{fig:resp_interp}), we can simulate patient-derived respiratory motion and match the  deformation of the lung anatomy with respect to published clinical lung respiratory displacement values~\cite{castillo_reference_2013, hofstad_intraoperative_2017, fried2023_IROS}.

%% file: Text/3_Discussion.tex
\section{Discussion}

In this work, we presented an extension to our prior work towards building a public dataset of anatomical environments for medical robots.
We built on our prior work by presenting three new environments of lungs anatomy that model the impact of respiratory deformation relevant to medical robots.
Given the challenges that deformation presents in medical procedures, medical robots operating in these settings need to reason and account for such deformation in the anatomy.
By providing these environments as a resource, we hope that our work can be used by the broad medical robotics community to help in designing and evaluating robot hardware and software solutions.
Specifically, we hope our work can alleviate the time-consuming process of gathering medical data and generating clinically meaningful environments, can foster reproducible research, and enable easier comparison between research works thereby driving the field forward as a whole.
We also envision that as this resource grows it will enable research teams to assess their robotic systems on a large and diverse set of anatomical environments that may be more representative of the heterogeneity in people.
Beyond serving as a resource for model-based approaches to medical robot development, these environments can be used as data-generation platforms for data-based approaches.
The files describing the anatomical environments can be integrated into custom simulators for medical robots~\cite{jianu2022cathsim, chentanez2009interactive, dreyfus2022simulation} and the simulated data can be used for learning tasks such as localization under respiratory deformation~\cite{fried2023_IROS}.
We hope to explore ways to integrate the dataset with simulators used by the broader robotics community~\cite{mittal2023orbit, coppeliaSim}.
Lastly, our virtual models can be a resource for 3D printing phantom models to aid in subsequent phases of robot evaluation.

In the future, we hope to increase the magnitude and diversity of medical environments in our dataset.
Additionally, deformable image registration in the lungs and respiratory motion modeling are active fields of research~\cite{liang2023orrn, shen2021accurate, duetschler2022synthetic} and we want to consider other registration algorithms for modeling respiratory motion.

%% file: Text/4_acknowledgements.tex
\section{Acknowledgements}

We thank the organizers of the Extraction of Airways from CT 2009 (EXACT09) study for their efforts in establishing the original study and for giving us permission to share the segmentations we generated using EXACT09 images.